\documentclass{article}

\usepackage{arxiv}

\usepackage[utf8]{inputenc} 
\usepackage[T1]{fontenc}    
\usepackage{hyperref}       
\usepackage{url}            
\usepackage{booktabs}       
\usepackage{amsfonts}       
\usepackage{nicefrac}       
\usepackage{microtype}      
\usepackage{lipsum}
\usepackage{amsmath} 
\usepackage{graphicx}
\graphicspath{ {./images/} }
\usepackage{subcaption} 

\title{Do We Need Reformer for Vision? \\ An Experimental Comparison with Vision Transformers}

\author{
ALI EL BELLAJ\thanks{Corresponding author: ae1057@msstate.edu} \\
  Department of Computer Science and Engineering\\
  Mississippi State University\\
  Starkville, MS 39759 \\
  \texttt{ae1057@msstate.edu} \\
   \And
 MOHAMMED-AMINE CHEDDADI \\
  Department of Computer Science and Engineering\\
 Mississippi State University\\
  Starkville, MS 39759 \\
  \texttt{mc3470@msstate.edu} \\
  \And
RHASSAN BERBER \\
  Department of Computer Science and Engineering\\
  Mississippi State University\\
  Starkville, MS 39759 \\
  \texttt{rb2400@msstate.edu} \\
}

\begin{document}
\maketitle
\begin{abstract}
Transformers have recently demonstrated strong performance in computer vision, with Vision Transformers (ViTs) leveraging self-attention to capture both low-level and high-level image features. However, standard ViTs remain computationally expensive, since global self-attention scales quadratically with the number of tokens, which limits their practicality for high-resolution inputs and resource-constrained settings.

In this work, we investigate the Reformer architecture as an alternative vision backbone. By combining patch-based tokenization with locality-sensitive hashing (LSH) attention, our model approximates global self-attention while reducing its theoretical time complexity from $\mathcal{O}(n^2)$ to $\mathcal{O}(n \log n)$ in the sequence length $n$. We evaluate the proposed Reformer-based vision model on CIFAR-10 to assess its behavior on small-scale datasets, on ImageNet-100 to study its accuracy--efficiency trade-off in a more realistic setting, and on a high-resolution medical imaging dataset to evaluate the model under longer token sequences.

While the Reformer achieves higher accuracy on CIFAR-10 compared to our ViT-style baseline, the ViT model consistently outperforms the Reformer in our experiments in terms of practical efficiency and end-to-end computation time across the larger and higher-resolution settings. These results suggest that, despite the theoretical advantages of LSH-based attention, meaningful computation gains require sequence lengths substantially longer than those produced by typical high-resolution images.
\end{abstract}


\section{Introduction}

Transformers first emerged in natural language processing (NLP) and quickly became a dominant architecture due to their strong ability to model long-range dependencies through self-attention. Since the original Transformer introduced by \cite{vaswani2023attentionneed}, this family of models has achieved state-of-the-art results in many NLP tasks, including machine translation, text summarization, and question answering. This success motivated researchers to explore whether the same attention-based principles could generalize beyond language into other modalities.

In computer vision, the introduction of Vision Transformers (ViTs) demonstrated that patch-based tokenization combined with self-attention can be a competitive alternative to convolutional neural networks (CNNs). By treating an image as a sequence of visual tokens, ViTs are able to capture both low-level patterns and higher-level semantic relationships in a unified framework. Compared to CNNs, self-attention offers a more direct mechanism to model global interactions across the image, which can be beneficial for recognition tasks where long-range context matters. However, similarly to NLP, transformer-based vision models are often computationally expensive and require large-scale training data to reach their full potential.

A core limitation of standard ViTs lies in the quadratic cost of global self-attention with respect to the sequence length. When images are processed at high resolution, the number of patches and tokens increases rapidly, leading to a significant growth in memory usage and training time. This makes vanilla self-attention less practical in resource-constrained settings and also raises concerns for domains such as medical imaging, where high-resolution inputs are important and computational budgets can be limited. Consequently, there is a growing interest in more efficient transformer variants that can maintain competitive accuracy while reducing computation.

Among the architectures proposed to address this issue, the Reformer introduced by \cite{kitaev2020reformerefficienttransformer} offers an attractive theoretical solution through locality-sensitive hashing (LSH) attention. Instead of computing attention across all token pairs, the Reformer groups tokens into buckets using hashing and restricts attention mostly within these buckets. This strategy reduces the theoretical attention complexity from $\mathcal{O}(n^2)$ to approximately $\mathcal{O}(n \log n)$ for sequence length $n$. While this idea has shown promise in language modeling, an important question remains: \emph{do such gains transfer meaningfully to vision tasks, where the token structure is derived from 2D patches rather than words?}

In this work, we explore this question by investigating a Reformer-based vision backbone. We combine standard patch embedding with LSH-based attention to build a Vision Reformer model, and we compare it against a ViT-style baseline under aligned training settings. We evaluate the models on CIFAR-10 to observe behavior on a small-scale dataset, on ImageNet-100 to examine the accuracy--efficiency trade-off in a more realistic recognition setting, and on a high-resolution diabetic retinopathy dataset to test performance under longer token sequences. Across these experiments, we aim to understand not only predictive accuracy but also end-to-end efficiency in practice.

The main contributions of this paper can be summarized as follows:
\begin{itemize}
    \item We adapt the Reformer architecture to vision by integrating patch tokenization with LSH attention.
    \item We provide a systematic comparison between a Reformer-based vision model and a ViT-style baseline under aligned experimental settings.
    \item We evaluate both models across small-scale, mid-scale, and high-resolution datasets to analyze accuracy and practical efficiency.
    \item We discuss the conditions under which LSH-based attention may or may not translate into real computation gains for vision tasks.
\end{itemize}

The remainder of this paper is organized as follows. Section 2 reviews related work on efficient Transformers for vision. Section 3 describes our proposed Vision Reformer architecture and experimental setup. Section 4 presents quantitative results and efficiency analyses. Section 5 discusses limitations and future research directions. Finally, Section 6 concludes the paper.

\section{Related Works}

Many efficiency ideas for Transformers were first developed in NLP and later adapted to vision. In this section, we discuss how these ideas influenced efficient Vision Transformer designs and motivate our investigation of Reformer-style LSH attention for image classification. However, before comparing such efficiency strategies, one must understand how attention is defined for images and why it can meaningfully replace convolution in early visual processing.

In vision, self-attention can be applied by treating pixels or patches as tokens, enabling each location to aggregate information from other spatial locations through learned similarity scores. Because images have inherent spatial structure, positional encoding is essential to avoid permutation ambiguity. \cite{cordonnier2020relationshipselfattentionconvolutionallayers} show that, with relative positional encoding, multi-head self-attention can simulate convolutional behavior. In particular, their Theorem~1 states that a multi-head self-attention layer with $N_h$ heads and a relative positional encoding of dimension $D_p \ge 3$ can express any convolutional layer with kernel size $\sqrt{N_h}\times\sqrt{N_h}$ and up to $\min(D_h, D_{out})$ output channels. This result helps justify why attention-based vision backbones can inherit strong local inductive biases while still allowing more global interactions when needed, supporting our exploration of LSH-based attention as an efficiency-oriented alternative.

To mitigate the quadratic cost of global attention, several vision transformers incorporate locality or hierarchical execution, methods that were firstly introduced in NLP. Swin-style models and their extensions introduce structured attention and improved token handling, aiming to reduce computation while maintaining competitive accuracy \cite{9874052,10643534}. Alternative attention formulations also explore replacing full quadratic interactions with structured approximations. Approaches such as attention via convolutional nearest neighbors highlight that attention-like behavior can be achieved through more constrained or efficient operators, potentially improving scalability depending on the target regime \cite{kang2025attentionconvolutionalnearestneighbors}. Recent survey work on Vision Transformers highlights that deploying ViTs efficiently on edge devices requires aggressive reductions in attention complexity, motivating alternative attention mechanisms such as structured, approximate, or hardware-friendly formulations \cite{saha2025visiontransformersedge}. A recent benchmark study compares over 30 efficient Vision Transformers under identical training and evaluation settings, revealing that ViT often remains Pareto-optimal across accuracy, speed, and memory trade-offs, while also emphasizing how architectural strategies such as sparse attention, token reduction, and hybrid attention affect practical efficiency \cite{nauen2024whichtransformertofavor}.

Beyond generic efficiency benchmarks, recent work has explored domain-specific and system-level optimizations. Mobile U-ViT introduces a lightweight U-shaped Vision Transformer tailored for medical image segmentation, combining large-kernel CNN-based hierarchical embeddings with a shallow transformer bottleneck to balance local inductive bias, long-range modeling, and strict computational constraints \cite{tang2025mobileuvit}. In parallel, TAFP-ViT targets efficiency from a hardware–software co-design perspective, proposing QKV computational fusion, adaptive token pruning, and specialized accelerator architectures to drastically reduce attention overhead and improve throughput and energy efficiency \cite{xu2025tafpvit}. Together, these works highlight complementary directions to attention approximation—namely domain-aware architectural design and system-level optimization—reinforcing the relevance of exploring Reformer-style LSH attention as a model-level efficiency mechanism that remains orthogonal to hardware or task-specific assumptions.

\section{Methods}

This section describes the Reformer architecture and our adaptation of it for vision classification. We then present our experimental design for evaluating accuracy--efficiency behavior across datasets with different token sequence lengths, and we detail the dataset-specific setups used in our comparison with a ViT baseline.

\subsection{Reformer Architecture}

Standard Vision Transformers rely on global self-attention, where each token attends to all other tokens in the sequence. Given a sequence length $n$, this requires computing an $n \times n$ attention matrix, resulting in $\mathcal{O}(n^2)$ time and memory complexity. While this is manageable for small images or coarse patching, it becomes expensive for higher resolutions where $n$ grows rapidly.

The Reformer was introduced as an efficiency-oriented alternative to full attention. Its central idea is to approximate global self-attention using locality-sensitive hashing (LSH). Instead of explicitly comparing each query against all keys, the Reformer uses a hashing function to group tokens into buckets such that tokens with similar representations are likely to fall into the same bucket. Attention is then computed primarily within each bucket (and sometimes across adjacent buckets), avoiding the full quadratic pairwise computation.

In practice, the Reformer achieves reduced complexity through two main mechanisms:

\begin{itemize}
    \item \textbf{LSH attention.} Tokens are projected using random matrices and assigned to hash buckets based on the sign or magnitude of projections. After hashing, tokens are sorted by bucket assignment. This sorting operation contributes to the $\mathcal{O}(n \log n)$ term, while attention within buckets is approximately linear in $n$ for fixed bucket sizes. The result is an attention mechanism that approximates global interactions with significantly reduced computational cost compared to standard self-attention.

    \item \textbf{Memory-oriented design.} The Reformer also introduces architectural choices that reduce memory overhead during training, such as reversible residual layers where intermediate activations are recovered from the output instead of being stored, and chunked feed-forward computation that reduces memory usage when computing large dimension vectors. These techniques reduce activation storage and help make training more feasible under long sequence lengths.
\end{itemize}

Together, these components provide a theoretical efficiency advantage over standard Transformer attention, particularly when $n$ becomes large.

\subsection{Vision Reformer via Patch Embedding}

To adapt the Reformer to vision, we follow the common ViT strategy of patch-based tokenization. An input image of size $H \times W$ is split into non-overlapping patches of size $P \times P$. Each patch is flattened and linearly projected into an embedding space of dimension $D$. This produces a sequence of $n$ tokens, where:
\begin{equation}
n = \left(\frac{H}{P}\right)\left(\frac{W}{P}\right).
\end{equation}

We then add positional information to preserve spatial structure and feed the resulting token sequence into a stack of Reformer encoder blocks using LSH attention. The final representation is aggregated using token-average pooling (or an equivalent global pooling strategy) and passed to a linear classification head.

This design preserves the core Reformer mechanism while making it compatible with standard image classification pipelines. Importantly, because the number of tokens is directly controlled by image resolution and patch size, this setup allows us to systematically test when the theoretical efficiency of $\mathcal{O}(n \log n)$ begins to offer practical benefits in vision.

\subsection{Experimental Design and Fair Comparison}

Our goal is to evaluate the accuracy--efficiency trade-off of the Vision Reformer under varying sequence lengths and to compare it against a standard ViT baseline. We perform supervised image classification experiments on three datasets representing different regimes of token count:

\begin{itemize}
    \item \textbf{Small-scale setting:} CIFAR-10, a 32 x 32 pixel color image dataset with 10 classes where the token sequence is relatively short.
    \item \textbf{Mid-scale setting:} ImageNet-100, a 224 x 224 dataset of 100 classes providing a more realistic recognition workload.
    \item \textbf{High-resolution setting:} a diabetic retinopathy dataset, where higher input resolution (600 x 600) leads to longer token sequences.
\end{itemize}

To ensure a fair comparison, both models are trained under aligned conditions:
\begin{itemize}
    \item \textbf{Same hardware:} all experiments are run on the same GPU and system configuration. \textit{an NVIDIA A100 40GB}
    \item \textbf{Matched model capacity:} we align the number of layers, embedding dimension, and number of attention heads between models.
    \item \textbf{Matched tokenization:} we use the same patch size and input resolution per dataset, ensuring both models operate on the same token sequence length $n$.
    \item \textbf{Matched training hyperparameters:} we use the same batch size (micro and effective), optimizer settings, learning rate schedule, and regularization where applicable.
\end{itemize}

To strengthen local feature extraction at early stages, we incorporate a lightweight convolutional stem in both the ViT baseline and the Vision Reformer. This stem applies small convolutional kernels before patch embedding, providing an explicit local inductive bias that helps capture edges, textures, and other short-range patterns. By enriching patch representations with local context, this design reduces the burden on the attention mechanism to learn purely local structure from scratch and can improve data efficiency and classification accuracy. Using the same convolutional front-end in both models also helps isolate the impact of the attention mechanism itself, making our Reformer--ViT comparison more controlled and fair.

We report both predictive and efficiency metrics. Predictive metrics include accuracy across datasets and additional macro-level metrics on the medical dataset. Efficiency metrics include end-to-end training time per epoch  and average GPU memory usage. This dual evaluation is critical because theoretical complexity reductions do not always translate into runtime gains under realistic vision sequence lengths.
We now summarize the training and preprocessing settings for each dataset.

\subsection{CIFAR-10}

The CIFAR-10 dataset consists of 60{,}000 color images of size $32 \times 32$ pixels. We use it to evaluate our models in a short-sequence regime, allowing us to observe whether LSH attention remains competitive in accuracy when the theoretical efficiency advantages of hashing are not expected to dominate runtime. We tokenize images using a patch size of $2 \times 2$, resulting in $16 \times 16 = 256$ tokens per image, and we set the depth of both architectures to 12 layers. For the Reformer model, we use a bucket size of 16 to encourage faster bucketed attention and to explicitly probe the behavior of the LSH mechanism in this small-token setting.

Both models are trained for 50 epochs using AdamW with weight decay for regularization to prevent underfitting. We apply Random  and Horizontal Flip augmentation to the dataset, as well as Random Crop to ensure the models properly fit the data. We employ a learning rate schedule with 5 warmup epochs followed by cosine annealing to stabilize early optimization and ensure consistent training dynamics across the Reformer and ViT baselines.

\subsection{ImageNet-100 }

We evaluate our models under a more computationally demanding setting using ImageNet-100, a subset of the ImageNet dataset containing 100 balanced classes. Compared to CIFAR-10, this benchmark provides higher-resolution inputs and a more diverse visual distribution. We resize images to $224 \times 224$ and, due to the increased dataset scale and training cost, we set the depth of both architectures to 10 layers. We use a patch size of $14 \times 14$, which yields $16 \times 16 = 256$ tokens per image. For the Reformer, we set the bucket size to 16 to reduce the cost of bucketed attention and to maintain a consistent attention configuration with our short-sequence experiments.

We do not further reduce the patch size to increase the number of tokens, since at this resolution smaller patches can introduce excessive local redundancy and may shift the models toward overly local representations, which can negatively affect accuracy. We apply the same augmentation strategy as in CIFAR-10, while using slightly lighter regularization, as the increased data diversity reduces the risk of underfitting in this setting.

\subsection{High-Resolution Diabetic Retinopathy}

To test the efficiency of the Reformer architecture under longer token sequences, we employ a high-resolution medical imaging dataset for diabetic retinopathy (DR). We choose this setting for two reasons. First, the dataset contains fundus images with a native resolution around $600 \times 600$, which naturally increases the token count in patch-based Transformers. Second, ViT-style models are increasingly used in healthcare applications, where fast, reliable, and computationally optimized inference and training pipelines are important. Evaluating a Reformer in this context is therefore practically relevant.

The dataset contains over 90{,}000 images labeled into five imbalanced classes reflecting clinical severity: 0) No DR, 1) Mild, 2) Moderate, 3) Severe, and 4) Proliferative DR. The majority of samples belong to the non-diabetic class, which reflects real-world screening conditions where positive cases are less frequent.

In this setting, we configure our models differently from the previous datasets to better stress-test attention efficiency. We resize images to $560 \times 560$, set the depth of both architectures to 8 layers, and use a patch size of $20 \times 20$. This results in $28 \times 28 = 784$ tokens per image, which is significantly higher than in our CIFAR-10 and ImageNet-100 experiments. We apply conservative medical-image augmentations, including horizontal flips, small rotations, and mild color jitter, in order to improve generalization without distorting clinically relevant features.

Both models are trained for 30 epochs with AdamW and a learning rate of $4 \times 10^{-4}$. Due to the longer sequence length and increased memory footprint, we use DeepSpeed to stabilize training and reduce GPU memory usage. Specifically, we enable mixed-precision training (FP16) and apply ZeRO optimization to shard optimizer states across devices, reducing the per-GPU memory cost. We also employ gradient accumulation to maintain a stable effective batch size while keeping a smaller micro-batch that fits into GPU memory. With gradient accumulation, the model performs multiple forward/backward passes before applying one optimizer update. If $\texttt{GRAD\_ACC\_STEPS} = k$, gradients are accumulated over $k$ micro-batches, and the effective batch size becomes:
\begin{equation}
    B_{\text{effective}} = B_{\text{micro}} \times \texttt{GRAD\_ACC\_STEPS}.
\end{equation}
This strategy is particularly important in our high-resolution setting, where increasing the micro-batch directly would lead to out-of-memory errors. Together, these DeepSpeed features allow us to evaluate the Reformer and ViT baselines under a longer-token, medically relevant regime while preserving aligned training conditions.

\section{Results}

In this section, we present the empirical evaluation of our Vision Reformer and the ViT baseline across three datasets with increasing token sequence lengths. Our goal is twofold: (1) to compare predictive performance under aligned capacity and training conditions, and (2) to assess whether the Reformer’s theoretical attention advantage translates into practical efficiency gains in vision. We therefore report both accuracy-oriented metrics and system-level measurements such as end-to-end training time per epoch and average GPU memory usage.

We organize the results by dataset to highlight how model behavior evolves from short to longer token regimes. We then provide a consolidated efficiency comparison to summarize the overall accuracy--runtime trade-off across all settings.

\subsection{CIFAR-10: Short-Sequence Regime}

\begin{figure}[h!]
    \centering
    
    \begin{subfigure}[b]{0.45\textwidth}
        \centering
        \includegraphics[width=\textwidth]{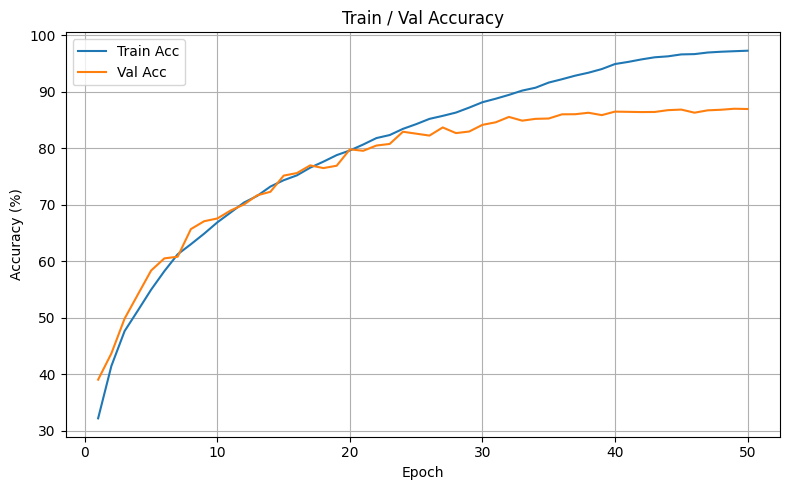} 
        \caption{Locality sensitive hashing attention}
        \label{fig:left_plot}
    \end{subfigure}
    \hfill 
    \begin{subfigure}[b]{0.45\textwidth}
        \centering
        \includegraphics[width=\textwidth]{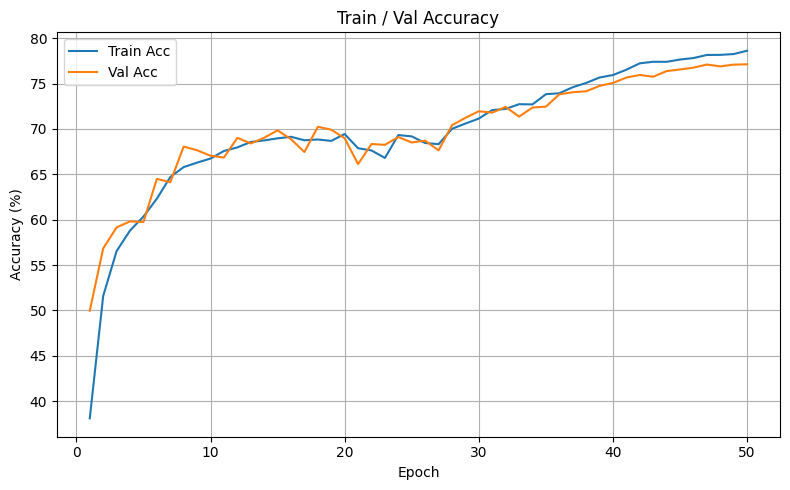}
        \caption{Self-attention}
        \label{fig:right_plot}
    \end{subfigure}
    
    \caption{Train/validation accuracy on CIFAR-10 for the Vision Reformer and ViT baseline under matched depth and tokenization (256 tokens).}
    \label{fig:cifar_acc}
\end{figure}

Figure~\ref{fig:cifar_acc} compares the training and validation accuracy of the Vision Reformer and the ViT baseline. The Reformer achieves a clearly higher validation accuracy in this short-sequence regime (86.95\%), suggesting that LSH-based attention can remain competitive---and even advantageous---for small-scale vision tasks. A plausible explanation is that the bucketed attention pattern, combined with our shared convolutional stem, introduces a structured inductive bias that helps optimization and acts as a mild form of regularization. Although the Reformer begins to show signs of overfitting around epoch 20, the validation curve continues to improve, indicating that the model is still learning useful decision boundaries rather than simply memorizing the training set. Table~\ref{TABLE 1} show all metrics from the reformer and ViT model.

Despite this accuracy advantage, the ViT remains substantially faster in terms of time per epoch (100 s for ViT against 165 s for reformer)  . This outcome is consistent with the fact that the Reformer’s efficiency benefits are primarily expected at longer sequence lengths. The original Reformer analysis shows that ``while regular attention becomes slower at longer sequence length, LSH attention speed remains flat'' \cite{kitaev2020reformerefficienttransformer}. In our CIFAR-10 setting, however, the token count is too small for this asymptotic advantage to dominate. The LSH pipeline introduces non-trivial constant overheads, including hashing via random projections, sorting and permuting tokens into buckets, and performing chunked attention within each bucket. Meanwhile, standard ViT attention at 256 tokens benefits from highly optimized dense GPU kernels, making the quadratic operation effectively cheap at this scale. As a result, the Reformer does not yet translate its theoretical $\mathcal{O}(n \log n)$ advantage into practical runtime gains in the short-sequence regime.

\begin{table}[ht]
    \centering
    \caption{Vision Transformers and Reformer architectures metrics.}
    \begin{tabular}{lccccc}
        \hline
        Model & Acc. & Prec. & Rec. & F1  & avg. time per epoch \\
        \hline 
        Reformer  & 86.95\%  & 86.9\% & 87\% & 86\%  & 165 s   \\
        ViT & 87.9\%  & 0.88\% & 0.88\% & 0.88\%  & 130 s   \\
        
        \hline
        \label{TABLE 1}
    \end{tabular}
    
\end{table}

\subsection{ImageNet-100: Mid-Scale Setting}

\begin{figure}[h!]
    \centering
    
    \begin{subfigure}[b]{0.45\textwidth}
        \centering
        \includegraphics[width=\textwidth]{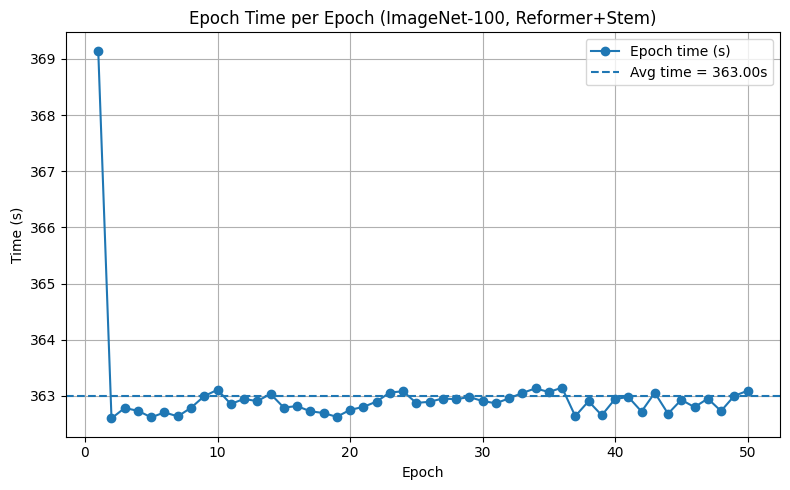} 
        \caption{Locality-sensitive hashing attention}
        \label{fig:left_plot}
    \end{subfigure}
    \hfill 
    \begin{subfigure}[b]{0.45\textwidth}
        \centering
        \includegraphics[width=\textwidth]{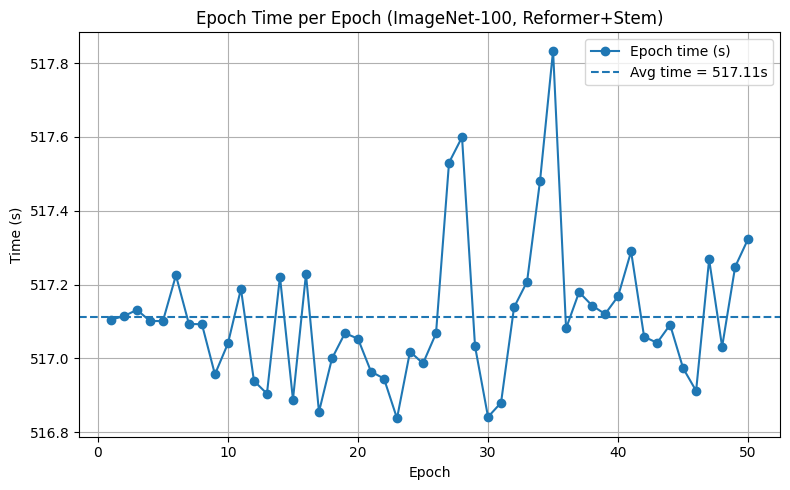}
        \caption{Self-attention}
        \label{fig:imagenet100_acc_time}
    \end{subfigure}
    
    \caption{average time per epoch on ImageNet-100 in seconds}
    \label{fig:imagenet_acc}
\end{figure}

On ImageNet-100, the gap between the two architectures becomes more apparent. The ViT baseline reaches a top-1 accuracy of 76.70\%, while the Vision Reformer attains 74.20\%, yielding a modest but consistent advantage for the ViT. The difference is more pronounced in terms of efficiency: the ViT requires on average 363\,s per epoch, compared to 517\,s for the Reformer (see Figure~\ref{fig:imagenet_acc}). This behavior reflects the fact that ImageNet-100 still operates in a moderate token regime (256 tokens), where dense self-attention remains highly optimized by existing GPU kernels. In contrast, the Reformer incurs additional overhead from hashing, sorting, and bucketed attention, without yet benefiting from the asymptotic gains of $\mathcal{O}(n \log n)$ attention. As a result, the Reformer cannot close the runtime gap and fails to translate its theoretical efficiency advantage into practical speedups at this scale.

\subsection{Diabetic Retinopathy: High-Resolution, Longer Tokens}

The diabetic retinopathy experiments reveal a different behavior once the token sequence becomes longer (784 tokens). In terms of efficiency, the gap between the two models almost disappears: the Reformer reaches an average epoch time of 1430\,s, while the ViT requires 1439\,s. This indicates that, at this token scale, the quadratic cost of dense self-attention begins to dominate for the ViT, allowing the Reformer’s more efficient attention pattern to partially offset its hashing and bucketing overhead. However, the Reformer is still not substantially faster, and the observed gain remains far from the ideal logarithmic regime suggested by its theoretical $\mathcal{O}(n \log n)$ complexity.

On the other hand, the ViT clearly outperforms the Reformer in predictive performance. The ViT achieves a top-1 accuracy of 74.56\%, whereas the Reformer reaches 70.96\%. This gap is consistent with the class imbalance of the dataset and is further reflected in the macro-precision and macro-recall scores reported in Table~\ref{TABLE 2}. The Reformer suffers more from the imbalance, yielding lower recall on the minority disease classes, while the ViT is better able to maintain balanced performance across the five severity levels. Overall, these results suggest that, in this high-resolution medical setting, the Reformer begins to approach the ViT in efficiency but still trails in accuracy, particularly in terms of sensitivity to underrepresented classes.
\begin{table}[ht]
    \centering
    \caption{Vision Transformers and Reformer architectures metrics.}
    \begin{tabular}{lcccccc}
        \hline
        Model & Acc. & Prec. & Rec. & F1  & auc & avg. time per epoch \\
        \hline 
        Reformer  & 70.96\%  & 63.7\% & 56.2\% & 58.8\%  & 88.7\% & 1430 s   \\
        ViT & 74.56\%  & 68.3\% & 62.6\% & 64.7\% & 90.9\% & 1439 s   \\
        
        \hline
        \label{TABLE 2}
    \end{tabular}
    
\end{table}

\section{Discussion}

Across all three datasets, the ViT baseline outperforms the Vision Reformer in terms of efficiency in the short- and mid-sequence regimes. On CIFAR-10 and ImageNet-100, where each image is represented by 256 tokens, dense self-attention remains highly optimized on current GPU hardware, and the additional overhead introduced by LSH attention makes the Reformer slower in practice. These results suggest that, for typical 2D image classification tasks with moderate token counts, standard ViTs are still a strong default choice.

The high-resolution diabetic retinopathy experiments provide a more nuanced picture. When we increase the token sequence to 784 tokens, the gap in efficiency between the two models almost disappears: the Reformer becomes slightly faster on average, although the difference is small. At the same time, the ViT maintains a clear accuracy advantage, especially on the minority disease classes. This indicates that increasing sequence length does move the Reformer closer to its theoretical efficiency regime, but not enough to produce large practical gains under our current configuration. Importantly, all experiments were conducted under a matched memory budget of roughly 12\,GB of GPU usage per model, with aligned depth, patch size, and optimizer hyperparameters, so the observed behavior reflects differences in the attention mechanism rather than differences in capacity or training setup.

One reason for the limited efficiency gains is that even high-resolution images do not generate sequences as long as those considered in the original Reformer work, where sequence lengths of many thousands of tokens were used. In our experiments, the maximum sequence length is 784 tokens, which is still relatively small compared to long-context NLP settings. At this scale, the $\mathcal{O}(n \log n)$ complexity of LSH attention does not yet clearly dominate the quadratic cost of standard self-attention. Instead, constant factors become important: hashing, sorting, and reordering tokens into buckets, as well as less optimized kernels for sparse or bucketed attention, introduce overheads that offset the theoretical complexity gains. In contrast, ViT attention is implemented using highly tuned dense matrix–multiplication kernels, which makes quadratic attention surprisingly efficient for moderate $n$.

From an accuracy point of view, our results also highlight a trade-off between approximate and exact attention. The Reformer’s LSH mechanism approximates global attention and may occasionally group dissimilar tokens or separate similar ones due to hashing collisions. While this approximation is acceptable or even beneficial in some language modeling scenarios, it can be more damaging in vision tasks where subtle global structure and fine-grained interactions are important, such as classifying rare disease stages in imbalanced medical datasets. The diabetic retinopathy results, where the Reformer suffers more on recall for minority classes, illustrate this sensitivity.

Overall, our study suggests that Reformer-style attention is not yet a drop-in replacement for standard ViT attention in 2D image classification with typical resolutions. However, the partial efficiency gains observed in the high-resolution medical setting indicate that LSH-based architectures could become more attractive in scenarios where sequence lengths are much higher than in our experiments. A natural direction is to consider modalities where the number of tokens can approach those of long-context NLP tasks, such as 3D medical volumes, video classification, or multi-frame temporal models, where each spatial–temporal patch contributes to a much longer sequence. In such settings, the Reformer’s asymptotic advantages may be more pronounced, and careful exploration of hybrid designs—combining convolutional stems, hierarchical tokenization, and LSH attention at selected stages—could yield more favorable accuracy–efficiency trade-offs.

Future work could also investigate more advanced strategies to mitigate class imbalance (e.g., focal loss or class-balanced reweighting) and to improve the robustness of LSH attention in vision, for example by learning data-dependent hashing schemes or combining LSH with token pruning. These extensions may help close the remaining accuracy gap while preserving or amplifying the emerging efficiency benefits observed at higher token counts.
\section{Conclusion}

In this work, we investigated the Reformer architecture as an efficient attention backbone for vision and compared it to a standard ViT under aligned settings on CIFAR-10, ImageNet-100, and a high-resolution diabetic retinopathy dataset. Our experiments show that, in all three cases, the ViT achieves higher accuracy except for CIFAR-10, and it remains faster in the short and mid-sequence regimes. Only in the high-resolution medical setting, with 784 tokens per image, does the Reformer begin to close the efficiency gap, but the gain is modest and does not compensate for the loss in accuracy.

These results suggest that, for typical 2D image classification tasks, Reformer-style LSH attention is not yet a practical replacement for standard ViT attention. However, the trends observed at longer token lengths indicate that Reformer-like architectures may become more attractive in domains with much longer sequences, such as 3D imaging or video, where the number of tokens can better match the regime envisioned in the original Reformer work. Exploring such settings, and designing hybrid models that combine strong inductive biases with efficient approximate attention, remains an important direction for future research.

\bibliography{references}
\bibliographystyle{apalike}

\end{document}